\documentclass{article}

\usepackage[preprint, nonatbib]{neurips_2023}

\usepackage{amsmath}
\usepackage{amssymb}
\usepackage{mathtools}
\usepackage{amsthm}
\usepackage{multirow}
\usepackage{enumitem}

\usepackage[utf8]{inputenc} 
\usepackage[T1]{fontenc}    
\usepackage{hyperref}       
\usepackage{url}            
\usepackage{booktabs}       
\usepackage{amsfonts}       
\usepackage{nicefrac}       
\usepackage{microtype}      
\usepackage{xcolor}         

\title{Joint fMRI Decoding and Encoding \\with Latent Embedding Alignment}

\author{%
  Xuelin Qian\thanks{Equal contribution} \\
  Fudan University \\
  \And
  Yikai Wang$^{*}$ \\
  Fudan University \\
  \And
  Yanwei Fu \\
  Fudan University \\
  \And
  Xinwei Sun \\
  Fudan University \\
  \And
  Jianfeng Feng \\
  Fudan University \\
  \And
  Xiangyang Xue \\
  Fudan University \\
}

\begin{document}

\maketitle

\begin{abstract}
The connection between brain activity and corresponding visual stimuli is crucial in comprehending the human brain. 
While deep generative models have exhibited advancement in recovering brain recordings by generating images conditioned on fMRI signals, accomplishing high-quality generation with consistent semantics continues to pose challenges. 
Moreover, the prediction of brain activity from visual stimuli remains a formidable undertaking. 
In this paper, we introduce a unified framework that addresses both fMRI decoding and encoding. 
Commencing with the establishment of two latent spaces capable of representing and reconstructing fMRI signals and visual images, respectively, we proceed to align the fMRI signals and visual images within the latent space, thereby enabling a bidirectional transformation between the two domains.
Our Latent Embedding Alignment (LEA) model concurrently recovers visual stimuli from fMRI signals and predicts brain activity from images within a unified framework. 
The performance of LEA surpasses that of existing methods on multiple benchmark fMRI decoding and encoding datasets. By integrating fMRI decoding and encoding, LEA offers a comprehensive solution for modeling the intricate relationship between brain activity and visual stimuli.

\end{abstract}

\section{Introduction} \label{sec:intro}
At each temporal instant, the human brain dynamically responds to visual stimuli conveyed through ocular reception~\cite{teng2019visual}, which can be indirectly quantified using functional Magnetic Resonance Imaging (fMRI). 
Identifying and categorizing distinct patterns of brain activity in reaction to visual stimuli is a crucial step in comprehending the enigma of the human brain. A significant approach to accomplish this is by inverse modeling, i.e., reconstructing the observed image from the fMRI signal~\cite{parthasarathy2017neural,horikawa2017generic}. Due to the intricate nature of images, the acquisition of pixel-level information poses challenges and is not invariably imperative. Consequently, researchers have primarily focused on decoding the semantic essence of images~\cite{horikawa2017generic,shen2019end, beliy2019voxels, gaziv2022self}. Conversely, fMRI encoding endeavors to predict the fMRI signal from visual stimuli.
Generally, the fMRI signal presents various inherent challenges:
\textbf{(1)} \emph{Redundancy}: Semantic information within the signal is sparsely distributed, with neighboring elements demonstrating high correlation, indicating redundant behavior of the fMRI signal~\cite{chang2019bold5000}.
\textbf{(2)} \emph{Instability}: The fMRI signal undergoes substantial influence from domain shifts, signifying that signals obtained from one individual or scanner may not be applicable for decoding fMRI signals from another individual or scanner~\cite{chen2022seeing}.
\textbf{(3)} \emph{Insufficiency}: In practical scenarios, the availability of image-signal pairs is limited, making it challenging to deploy prevalent deep learning methods that heavily rely on large training sets for fMRI comprehension.

\begin{figure}
\begin{centering}
\includegraphics[width=1.0\textwidth]{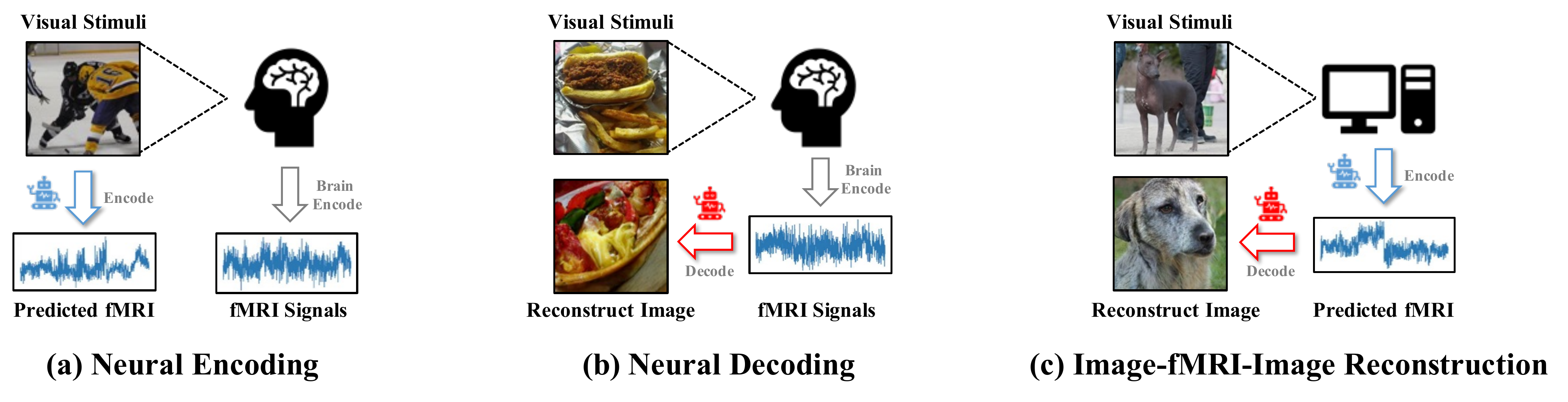}
\vspace{-0.15in}
\caption{
Our proposed LEA efficiently enables (a) neural encoding to estimate the brain activity from visual stimuli, (b) neural decoding to recover the visual stimuli from recorded fMRI signals, and (c) image-fMRI-image reconstruction to ensure the reliability of neural decoding and encoding. \label{fig:intro}
}

\end{centering}
\vspace{-0.15in}
\end{figure}

Recent works~\cite{ozcelik2022reconstruction, Matteo2022semantic, chen2022seeing, liu2023brainclip, du2023decoding, ozcelik2023brain} attempted to decode fMRI signals based on pre-trained generative models like Instance-Conditional GAN~\cite{casanova2021instance}, diffusion models~\cite{ho2020denoising}, masked autoencoders~\cite{he2022masked}, CLIP~\cite{radford2021learning}, to name a few. 
Despite achieving impressive results in high-fidelity generation, these methods encounter several inherent challenges:
(1) while pre-trained generative models prove advantageous in generating images of exceptional quality, ensuring semantic consistency with fMRI signals remains a persistent challenge.
(2) These models demonstrate the ability to generate high-quality images even when presented with random noise masquerading as fake fMRI signals, thereby raising concerns regarding their reliability, particularly in the context of open-vocabulary visual stimuli.
Furthermore, only a handful of methodologies jointly addressing fMRI decoding and encoding tasks.

In this paper, we present a pioneering framework that tackles the challenges of fMRI signals through the joint tasks of fMRI decoding and encoding. Our framework has several key components.
(1) First, we introduce an encoder-decoder architecture designed specifically for fMRI signals and images. By utilizing this architecture, we effectively learn a dense and compact latent representation space for each modality. This approach directly addresses the issue of signal redundancy in fMRI data, offering promising solutions.
(2) Additionally, our framework demonstrates the advantage of training the encoder-decoder architectures independently for fMRI signals and images. This eliminates the necessity of paired fMRI-image data, thereby circumventing the insufficiency problem associated with such datasets. Our method provides a practical solution that overcomes the limitations of conventional approaches.
(3) Despite the inherent instability of fMRI signals across different individuals, our framework successfully compresses these signals into a unified latent representation space using a shared encoder. 
(4) Inspired by the principles of self-supervised training~\cite{vincent2008extracting, he2020momentum, he2022masked}, we leverage this methodology to train our encoder-decoder architecture. This approach enables the construction of a dense and compact latent representation space without the need for extra supervision. 

Furthermore, we propose a lightweight alignment module to link the two latent spaces that can be learned with relatively small fMRI-image pairs.
Such an alignment module is individual-dependent to alleviate the instability issue of the fMRI signals.
The alignment of different signals has been widely studied in vision tasks like image-text~\cite{radford2021learning,jia2021scaling,mahajan2018exploring} and image-audio~\cite{arandjelovic2017look,morgado2021audio,owens2018audio,patrick2020multi}. 
These approaches have discovered that when signals from different modalities are well-encoded, they can be aligned in the same representation space.
However, we do not have a large fMRI-image paired dataset to support the end-to-end training of such an alignment model.
Hence, we instead try to link fMRI and image representations that are separately learned.
Specifically, inspired by the linear relation discovered between text- and vision-only models~\cite{merullo2023linearly}, we also adopt a linear regression model from one latent space to the other to learn the relationship between two latent spaces.
Empirical results suggest that such linear relation also exist in fMRI- and vision-only models.
Furthermore, such a lightweight linear model can be learned with small fMRI-image pairs, which avoids the requirement of large-scale paired dataset for alignment model and make it possible to learn individual-specific models.
Utilizing the alignment module within LEA, we can assess the reliability of both the neural encoding and decoding models through the image-fMRI-image reconstruction task. 
By initially estimating the fMRI signals from the image and subsequently reconstructing the visual stimuli, the ability to reconstruct semantically consistent images serves as an indicator of their reliability.

Drawing from these analyses, we have devised the Latent Embedding Alignment (LEA) framework tailored for fMRI decoding and encoding. 
LEA incorporates two distinct encoder-decoder architectures for handling fMRI signals and images, respectively. 
We introduce an alignment module to facilitate the transformation between the latent representation spaces of fMRI signals and images.
As depicted in Figure~\ref{fig:intro}, when recovering visual stimuli from fMRI signals, we encode the fMRI signal, transform it into a latent image embedding, and subsequently decode it to generate images. 
Conversely, in the context of predicting brain activities, we encode the images, transform them into latent fMRI embeddings, and then decode them to fMRI signals. 
Notably, LEA seamlessly supports image-fMRI-image reconstruction, thereby ensuring the reliability of both neural encoding and decoding models.
Through extensive experimentation on various benchmark datasets, we demonstrate that LEA exhibits both efficiency and effectiveness in fMRI decoding and encoding.

\textbf{Contribution}. Our contributions can be succinctly summarized as follows:
\begin{itemize}[leftmargin=*,itemsep=0pt,topsep=0pt,parsep=0pt]
\item We facilitate the training of encoder-decoder architectures to acquire latent spaces specifically tailored for fMRI signals and images.
\item We establish a connection between these two latent spaces, thus enabling fMRI decoding and encoding within a unified framework dubbed as LEA.
\item Through rigorous validation on multiple benchmark datasets, we demonstrate the superior performance of LEA, highlighting its efficacy and advancements.
\end{itemize}

\section{Related Work} \label{sec:related-work}

\textbf{Neural Coding}. It is the study including both neural encoding and decoding. 
For neural encoding, it refers to learn the map from visual stimulus to fMRI signals. The early works \cite {yamins2013hierarchical,yamins2014performance,gucclu2015deep,yamins2016using} mainly use convolutional neural networks to encode semantics from visual stimuli. With the renaissance of neural language models, recent studies leverage BERT \cite{devlin2018bert}, Transformer \cite{Vaswani2017Attention} and GPT-2 \cite{radford2019language} successfully predict fMRI responses from stimuli, including images or even words and sentences.
On the contrary, neural decoding focuses on the reverse map, from the brain activity to stimulus.
Previous studies ~\cite{mozafari2020reconstructing, ozcelik2022reconstruction} developed regression models specifically designed to extract valuable information from the fMRI signal.
Ozcelik \textit{et al.} \cite{ozcelik2022reconstruction} utilized a pre-trained Instance-Conditional GAN model \cite{casanova2021instance} to reconstruct images by decoding latent variables from fMRI data. Ferrante \textit{et al.} \cite{Matteo2022semantic} employed pre-trained latent diffusion models \cite{ho2020denoising} to generate stimuli images by mapping fMRI signals to visual features. Chen \textit{et al.} \cite{chen2022seeing} introduced a self-supervised sparse masked modeling strategy to encode fMRI data into latent embeddings and fine-tuned latent diffusion models with double conditioning. In this paper, we design an unified framework to realize both nerual encoding and decoding efficiently.

\textbf{Multi-Modality Alignment.} It aims to link between different modalities, for example image-text alignment~\cite{radford2021learning,jia2021scaling,mahajan2018exploring} and image-audio alignment~\cite{arandjelovic2017look,morgado2021audio,owens2018audio,patrick2020multi}. 
Some work further attempts to align several modalities in the same latent representation space~\cite{girdhar2023imagebind}.
Most works in this direction based on the CLIP model~\cite{radford2021learning} which is trained on large-scale image-text pairs to learn a single latent space where the distance between correlated image-text pairs are minimized. For example, Liu \textit{et al.} \cite{liu2023brainclip} bridged the modality gap by leveraging CLIP's cross-modal generalization ability to address the limitation of insufficient fMRI-image paired data. Similarly, Du \textit{et al.} \cite{du2023decoding} used multi-modal deep generative models to capture the relationships between brain, visual, and linguistic features. Ozcelik \textit{et al.} \cite{ozcelik2023brain} combined these techniques to predict multi-modal features from fMRI signals and generate reconstructed images using a latent diffusion model \cite{rombach2022high}.

\section{Methodology} \label{sec:method}

\textbf{Problem Setup}.
Given fMRI signal $\mathbf{f}$ recorded from brain activity and the corresponding visual stimuli, image $\mathbf{I} \in \mathbb{R}^{H\times W \times 3}$
, the purpose of this paper is to learn a unified framework that can perform both \textbf{neural decoding} task that recovering the observed image from fMRI signal as well as \textbf{neural encoding} task that predicting the brain activity from the image.

For decoding task, ideally, the reconstructed images $\hat{\mathbf{I}}$ should be the same as the real one. Considering the peculiarity of biological mechanisms, in which an individual is affected by individual memory or attention, the brain activity stimulated by the same image can be quite different. Therefore, we follow \cite{mozafari2020reconstructing,ozcelik2022reconstruction} to require the recovered images $\hat{\mathbf{I}}$ to be \textit{semantically consistent} with $\mathbf{I}$.
Thus we focus on the semantic information in the image instead of pixel-level structure. 

For encoding task, it is difficult to precisely predict element-wise fMRI signals as itself is noisy and redundancy.
Thus, in this paper we focus on the prediction of the tendency of fMRI signals.
Specifically, we expect the estimated fMRI signals to have a correspondence with the real fMRI signals which can be used to indicate whether a ROI region is activated or not.
To this end, our target is to produce estimated fMRI signals that are linearly related to the real fMRI signals.
Such relationship can be checked via Pearsonr correlation.

\begin{figure*}[t]
\begin{center}
\centerline{\includegraphics[width=0.95\textwidth]{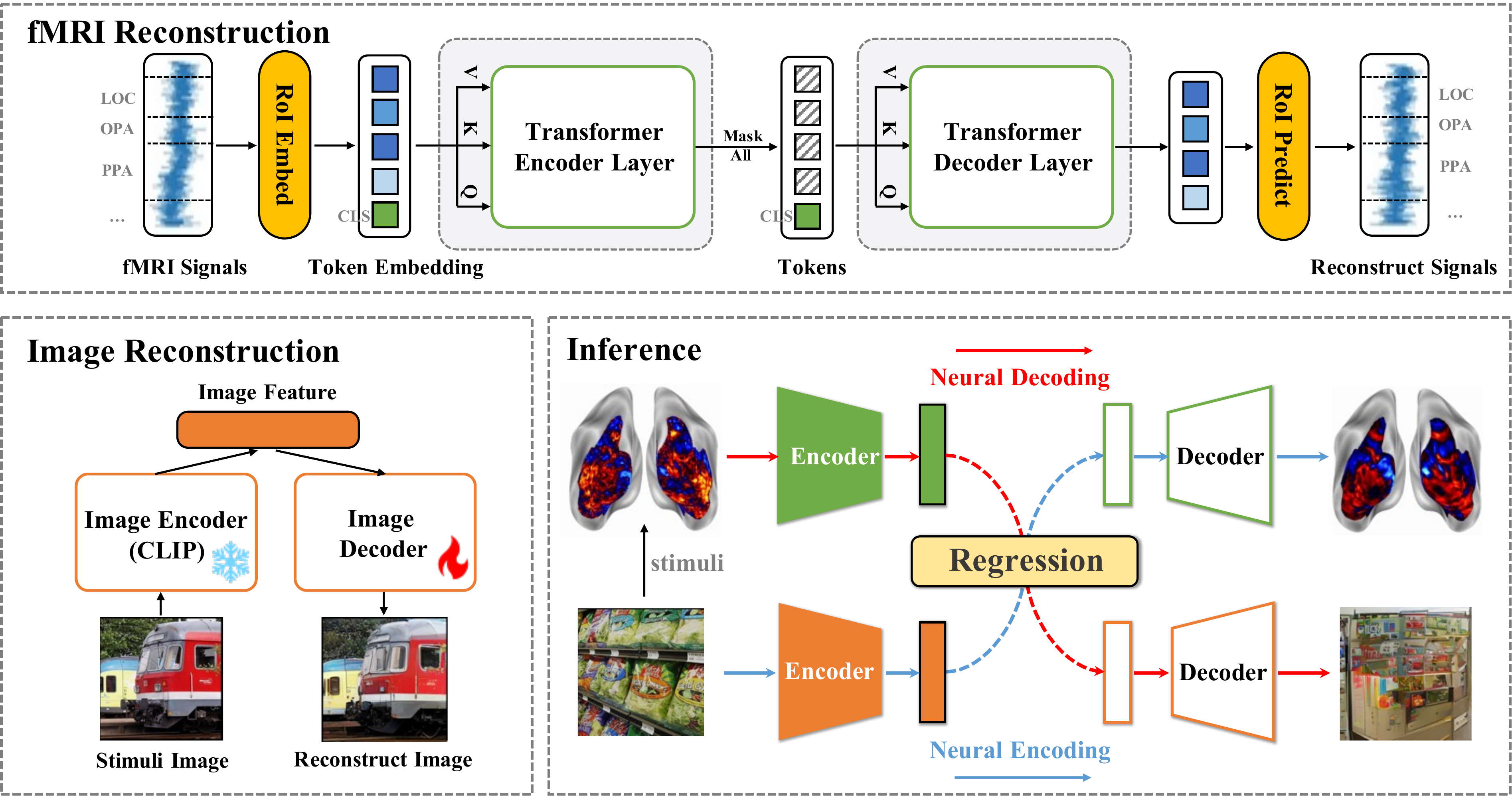}}
\vspace{-0.1in}
\caption{Illustration of LEA. 
We learn a latent representation space via encoder-decoder architecture designed specifically for fMRI signals and images.
Then a lightweight alignment module is proposed to link the two latent spaces, enabling joint fMRI encoding and decoding.
}
\label{fig:framework}
\end{center}
\vspace{-0.15in}
\end{figure*}

\subsection{Latent Space Construction}

\subsubsection{fMRI Latent Space Construction}
The fMRI signals indirectly records neural activity in the visual cortex of the brain by measuring the blood-oxygen-level-dependent (BOLD) signals, which is usually represented as data of 3D voxels, covering visual regions, like EarlyVis, LOC, OPA, PPA, and RSC \cite{Horikawa2015Generic}. 
When observing different images, fMRI signals will have different patterns.
For the same individual in different running experiments, the fMRI signals show similar behavior.
The biological principles indicate that neighboring voxels in the brain's visual cortex often show a similar intensity of stimulus-response \cite{Ugurbil2013Pushing}, leading to high redundancy of fMRI signals. 

To explore the potential responding patterns according to the visual stimuli and build the reverse mapping for semantic encoding, it is thus necessary to focus on the relationships among different regions and simultaneously perceive the activity changes from long-range voxels. Inspired by the advantage of long-range receptive fields from transformer layers, we follow MinD-Vis~\cite{chen2022seeing} to adopt the architecture of masked autoencoder~\cite{he2022masked} as the encoder-decoder model for fMRI signals.

However, different from \cite{chen2022seeing} that equally splits the fMRI signals into patches of size 16, we instead maintain the voxel structure with ROI regions.
Specifically, we first divide the whole fMRI signal vector (after pre-processing) into different ROI regions, where each of them may have different dimensions.
Thus, a ROI-dependent pre-processing is required.
To this end, we introduce a RoI Embed layer. Concretely, given a specific ROI region $i$, assume the corresponding fMRI signal is of size $N_i \times 1$, which indicate $N_i$ voxels and 1 dimension of BOLD response.
Such a fMRI signal is first feed into a convolution layer with 32 kernels to extract multi-head features as $N_i \times 32$.
Then we adopt a fully-connected layer to project the different length of fMRI signal into a unified dimension of 1024, leading to the final fMRI embedding as $32\times 1024$.
The input to the encoder is thus the concatenation of fMRI embeddings of different ROI regions.

The original MAE adopts the patch-based reconstruction to mask a ratio of the patches and then predict the masked patches.
While this is beneficial for learning a local representation, it is not straightforward if we want to learn a global latent code to represent the whole fMRI signal.
To this end, we instead only preserve the global [CLS] token of the encoded fMRI singals, and mask all the patch-based latent code and let the decoder to learn to reconstruct fMRI signals from the learned [CLS] token.
Such a strategy is helpful to learn a dense fMRI representation in the [CLS] token and the learned decoder can be directly used as a generator to generate fMRI signals from the estimated latent fMRI code.

Finally, the output of the decoder is post-processed by a RoI Project layer, which has the symmetric structures of the RoI Embed layer, to reverse the pre-processing on the fMRI signals such that the length is identical to the original fMRI signals. The overall encoder-decoder architecture is trained to minimize the L2 reconstruction loss of the fMRI signals.
\begin{equation}
    \mathcal{L}_{fmri}= \frac{1}{M}\sum_{i=1}^{M}||\mathbf{f}_{i} - \hat{\mathbf{f}}_{i}||_{2}^{2}
\end{equation}
\noindent where $\mathbf{f}_{i}$ and $\hat{\mathbf{f}}_{i}$ indicate the ground-truth and reconstructed fMRI signals; $M$ means the totally number of training fMRI data.

\subsubsection{Image Latent Space Construction}
Image-based latent space construction has been widely investigated.
As our target is to recover the semantic content of fMRI signals, we adopt the pre-trained CLIP~\cite{radford2021learning} visual encoder and the yielding CLIP latent space as our image latent space. 
CLIP is trained on large-scale image-text paired, thus the learned latent space is semantic-rich and is beneficial for the semantic recovery of fMRI signals.

However, CLIP is a encoder-only model, and we cannot to directly generate images from the image latent space.
To generate images from the image latent space, we adopt MaskGIT~\cite{Chang2022maskgit} as the decoder in this paper to produce reconstruction of the image, but other generative models are also acceptable. Specifically, we freeze the CLIP visual encoder and fine-tune the class-conditional MaskGIT to learn a latent-conditional generative decoder. We refer readers to \cite{Chang2022maskgit} for details of objective functions.

\subsection{Latent Embedding Alignment}
After the construction of fMRI and image latent spaces, the last step is to alignment the two spaces such that we can jointly perform fMRI decoding and encoding tasks.
Different from previous work \cite {du2023decoding,liu2023brainclip} that leverage additional modalities to help bridge the gap between fMRI and image, we advocate that well-represented latent features of both fMRI and image data can be connected via a simple linear model. Furthermore, considering that 
fMRI signals suffer from the instability issue, the latent fMRI embedding of different individuals may have different patterns.
we thereby learn an individual-specific linear model to map the personalized fMRI embedding to the image embedding and vice versa.
As the two latent embeddings are both high dimensional, we add L2 regularization to constrain the learning space (\textit{i.e.}, ridge regression). 

After constructing the two latent spaces and learning the latent embedding alignment module, we concurrently perform fMRI decoding and encoding, all within the confines of a unified framework.
For fMRI decoding task, we first encode the fMRI signal in the learned latent fMRI space, then we adopt the individual-specific linear model to project the latent fMRI code to the latent visual feature, and such visual feature is decoded in the fine-tuned class-conditional MaskGIT to perform recovery of the visual stimuli.
For fMRI encoding task, we first encode the CLIP visual feature, and then project it to the latent fMRI space via the individual-specific linear model.
The projected latent fMRI code is decoded to estimate the real fMRI signals.

\section{Experiment} \label{sec:experiment}


\noindent \textbf{Datasets.} 
We validate our LEA on Brain, Object, Landscape Dataset (BOLD5000)~\cite{chang2019bold5000} and Generic Object Decoding Dataset (GOD)~\cite{horikawa2017generic}. 
(1) BOLD5000 is a large-scale fMRI-image dataset, which consists of a total of $5,254$ fMRI-image stimuli trials. 
Four subjects are involved in this experiment. 
Among these natural images, $1000$, $2000$ and $1916$ images are from Scene, COCO, and ImgeNet datasets, respectively. 
Following the protocol in~\cite{chen2022seeing}, we choose $4,803$ images presented on a single trial for training, and the remaining $113$ images for testing. We use the pre-defined five ROI regions (\textit{e.g.},EV, LOC, OPA, PPA, RSC) for our RoI Embed layer. 
(2) GOD involves 5 subjects, each of which is asked to see 1250 images from 200 categories, resulting in a total of 1250 fMRI-image pairs for the study. We split the training set with 1200 pairs from 150 categories, and the other non-overlapping 50 classes are used for testing.
The ROI regions we used contain V1, V2, V3, V4, FFA, PPA, LOC and HVC.

\noindent \textbf{Metrics.} 
For the neural decoding task, we adopt the following metrics to provide comprehensive and in-depth comparisons with other competitors. 
(1) N-way Classification Accuracy (Acc)~\cite{gaziv2022self} are adopted to measure the semantic correctness of the generated samples. We calculate the average accuracy of top-1 for 1000 trials. 
(2) Fr\'echet Inception Distance (FID)~\cite{heusel2017gans} measures the quality of generated images. We use features with 64 dimensions extracted by the inception-v3 model to compute the similarity with ground-truth testing images.
(3) CLIP correlation and distance to measure the semantic consistent of the generated images. 
The CLIP distance is the average of the CLIP similarity among all the fMRI-image pairs, while the CLIP correlation is calculated to measure the correspondence of the positive fMRI-image pair compared with other .

For the neural encoding task, we adopt the Pearsonr metric to measure the similarity between estimated fMRI signals and the ground-truth.
Given the $N\times L$ matric of $N$ fMRI signals of vertexes $L$, the pearsonr correlation is calculated on each vertexes between $N$ estimated fMRI signals and the ground-truth and then averaged among different vertexes.

\noindent \textbf{Implementation.} 
All experiments are implemented with the PyTorch toolkit. For fMRI reconstruction model, it is initialized with pre-trained weights from HCP datasets \cite{chen2022seeing}. We further finetune it with both fMRI training and testing sets for each individual. Notably, using fMRI testing data is a common practice as in \cite{chen2022seeing,beliy2019voxels}, and we do not access any paired fMRI-image testing data when learning linear models. The depth of the encoder and the decoder is 24/8 with 1024/512 dimensions. The number of both multi-heads is set to 16. 
To finetune both models, AdamW optimizer is applied with $\beta_{1}=0.9$, $\beta_{1}=0.95$ and batch size $8$. The initial learning rate is defined as 5e-5 with 0.01 weight decay. We linear decrease the learning rate until it reaches to the minimal rate. The total number of training iterations for the fMRI reconstruction model and the image reconstruction model is 100k and 300k.

\subsection{Main Results}

\begin{table}
\centering{
\small{
\caption{Top-1 accuracy of zero-shot classification on GOD dataset.\label{tab:fMRI-to-Image}}
\setlength{\tabcolsep}{2.5mm}{
\begin{tabular}{lccccccc}
\toprule
\multicolumn{1}{c}{{\textsc{Method}}} & {\textsc{Modality}} & \textsc{Sbj-1} & \textsc{Sbj-2} & \textsc{Sbj-3} & \textsc{Sbj-4} & \textsc{Sbj-5} & {\textsc{Average}} \\
\midrule
\midrule
CADA-VAE \cite{schonfeld2019generalized} & V\&T & 6.31 & 6.45 & 17.74 & 12.17 & 7.45 & 10.02  \\
MVAE \cite{wu2018multimodal} & V\&T & 5.77 & 5.40 & 17.11 & 14.02 & 7.89 & 10.04  \\
MMVAE \cite{shi2019variational} & V\&T & 6.63 & 6.60 & 22.11 & 14.54 & 8.53 & 11.68  \\
MoPoE-VAE \cite{sutter2021generalized} & V\&T & 8.54 & 8.34 & 22.68 & 14.57 & 10.45 & 12.92  \\
BrainCLIP-Linear \cite{liu2023brainclip} & V\&T & 10.00 & 12.00 & 18.00 & 12.00 & 12.00 & 12.80 \\
\midrule
CADA-VAE \cite{schonfeld2019generalized} & V & 5.66 & 6.01 & 16.51 & 9.17 & 6.01 & 8.67  \\
MVAE \cite{wu2018multimodal} & V & 5.30 & 5.21 & 14.13 & 8.03 & 5.44 & 7.62  \\
MMVAE \cite{shi2019variational} & V & 5.41 & 5.39 & 13.76 & 10.62 & 5.02 & 8.04  \\
MoPoE-VAE \cite{sutter2021generalized} & V & 5.20 & 7.42 & 14.05 & 9.25 & 6.37 & 8.46  \\
BraVL \cite{du2023decoding} & V & 8.91 & 8.51 & \textbf{18.17} & 14.20 & 11.02 & 12.16  \\
BrainCLIP \cite{liu2023brainclip} & V & 6.00 & 16.00 & 16.00 & 14.00 & \textbf{14.00} & 13.20 \\
\midrule
LEA & V & \textbf{10.00} & \textbf{18.00} & 12.00 & \textbf{16.00} & 12.00 & \textbf{13.60}    \\
\bottomrule
\end{tabular}}}}
\vspace{-0.15in}
\end{table}

\begin{table}
\centering{
\small{
\caption{fMRI encoding comparison on Pearsonr metric (The higher the better). \label{tab:Image-to-fMRI}}
\setlength{\tabcolsep}{1.1mm}{
\begin{tabular}{lccccccccccc}
\toprule
\multicolumn{1}{c}{\multirow{2}{*}{\textsc{Method}}} & 
\multicolumn{6}{c}{\textsc{GOD}} & \multicolumn{5}{c}{\textsc{BOLD5000}} \\
\cmidrule{2-12}
& {Sbj-1} & {Sbj-2} & {Sbj-3} & {Sbj-4} & {Sbj-5} & Avg. &  {CSI-1} & {CSI-2} & {CSI-3} & {CSI-4} & Avg. \\
\midrule
\midrule
CLIP + MinD-Vis \cite{chen2022seeing} & 5.21 & 12.05 & 15.47 & 10.63 & 10.55 & 10.78 & 30.58 & 19.78 & 16.48 & 19.96 & 21.70\\
CLIP + Regressor \cite{gifford2023algonauts} & 6.65 & \textbf{16.22} & 18.58 & 12.82 & 14.38 & 13.73 & 32.07 & 20.58 & 20.46 & 21.18 & 23.57 \\
\hline
LEA & \textbf{8.43} & 15.51 & \textbf{21.20} & \textbf{19.08} & \textbf{15.07} & \textbf{15.86} & \textbf{34.32} & \textbf{26.74} & \textbf{27.74} & \textbf{24.44} & \textbf{28.19}  \\
\bottomrule
\end{tabular}}}}
\end{table}

\textbf{Zero-shot classification}.
As we inherit the latent space of original CLIP model, our LEA enjoys an open-vocabulary recognition of fMRI signals.
To validate this, we conduct experiments on zero-shot fMRI signal classification tasks to recover the most similar category of the fMRI signals.
Specifically, we design a set of candidate class names via several text templates.
These classes are encoded via the CLIP text encoder.
Then we can direct calculate the similarity between class embeddings and fMRI latent embeddings learned via LEA.

We conduct experiments on the test set of GOD, which contains 50 visual stimuli from 50 categories that are unseen during training.
We report the top-1 classification accuracy (with the chance levels of 2\%) of several competitors in Table~\ref{tab:fMRI-to-Image}.
We divide the competitors into two groups based on their supervision modality, where V indicates visual supervision only and V\&T indicates bothe visual and textual supervision.
Our LEA trained with visual supervision beats all competitors, indicating the superiority of our bi-directional transformation.

\textbf{Joint fMRI encoding and decoding}.
With two latent spaces for fMRI signal and image, Our LEA can not only recover the visual stimuli but also predict fMRI signal based on visual image. 
As shown in Table~\ref{tab:Image-to-fMRI}, we conduct experiments to validate the estimation capability of our LEA. 
We construct two competitors to show the effectiveness of LEA.
The `CLIP+Regressor' means that we apply liner regression to directly generate fMRI signal based over the image feature from CLIP model. 
However, such method causes misalignment of features during reconstruction without the fMRI latent space. 
The `CLIP+MinD-Vis' which utilizes the transformer decoder to predict fMRI signal according to the image feature from CLIP model.
Such solution is too rough and not conducive to fine-grained feature alignment and generation. 
Our LEA achieves the best Pearsonr score, which indicates that the alignment between latent embedding spaces are more effective for fMRI prediction.

\begin{table} 
\centering{
\small{
\caption{fMRI decoding performance comparison on GOD dataset. \label{tab:god}}
\setlength{\tabcolsep}{2.7mm}{
\begin{tabular}{llcccccc}
\toprule
{\textsc{Metrics}} & {\textsc{Methods}} & \textsc{Sbj-1} & \textsc{Sbj-2} & \textsc{Sbj-3} & \textsc{Sbj-4} & \textsc{Sbj-5}  & \textsc{Average} \\
\midrule
\midrule
\multirow{3}{*}{FID $\downarrow$} & Gaziv \textit{et al.} \cite{gaziv2022self} & 7.67 & 2.67 & 2.51 & 8.93 & 2.77 & 4.91 \\
& MinD-Vis \cite{chen2022seeing} & 1.97 & 1.63 & 1.68 & 1.77 & 2.33 & 1.88 \\
& \textit{Ours} & \textbf{1.45} & \textbf{1.43} & \textbf{1.52} & \textbf{1.48} & \textbf{1.25} & \textbf{1.23} \\
\midrule

\multirow{3}{*}{CLIP Corr. $\uparrow$} & Gaziv \textit{et al.} \cite{gaziv2022self} & 58.57 & 62.00 & 67.59 & 62.89 & 61.14 & 62.44 \\
& MinD-Vis \cite{chen2022seeing} & 66.69 & 77.02 & \textbf{83.78} & 74.78 & \textbf{80.04} & 76.46 \\
& \textit{Ours} & \textbf{73.96} & \textbf{78.20} & 80.94 & \textbf{78.73} & 70.86 & \textbf{76.54} \\
\midrule

\multirow{3}{*}{CLIP Dist. $\uparrow$} & Gaziv \textit{et al.} \cite{gaziv2022self} & 0.31 & 0.34 & 0.34 & 0.32 & 0.32 & 0.33 \\
& MinD-Vis \cite{chen2022seeing} & 0.33 & 0.35 & \textbf{0.43} & 0.38 & \textbf{0.39} & 0.38 \\
& \textit{Ours} & \textbf{0.38} & \textbf{0.40} & 0.41 & \textbf{0.40} & 0.37 & \textbf{0.39} \\
\midrule

\multirow{3}{*}{Accuracy $\uparrow$} & Gaziv \textit{et al.} \cite{gaziv2022self} & 1.88 & 3.79 & 12.99 & 8.45 & 6.26 & 6.67 \\
& MinD-Vis \cite{chen2022seeing} & 9.10 & 15.91 & \textbf{27.44} & 15.81 & \textbf{14.28} & 16.51 \\
& \textit{Ours} & \textbf{11.18} & \textbf{18.62} & 20.45 & \textbf{20.04} & 13.15 & \textbf{16.69} \\
\bottomrule
\end{tabular}}}}
\end{table}

\begin{table} 
\centering{
\small{
\caption{fMRI decoding performance comparison on BOLD5000 dataset. \label{tab:bold}}
\setlength{\tabcolsep}{3.2mm}{
\begin{tabular}{llccccc}
\toprule
{\textsc{Metrics}} & {\textsc{Methods}} & \textsc{CSI-1} & \textsc{CSI-2} & \textsc{CSI-3} & \textsc{CSI-4} & \textsc{Average} \\
\midrule
\midrule
\multirow{2}{*}{FID $\downarrow$}
& MinD-Vis \cite{chen2022seeing} & \textbf{1.20} & 1.90 & \textbf{1.40} & 1.32 & 1.46  \\
& \textit{Ours} & 1.31 & \textbf{1.53} & 1.48 & \textbf{1.19} & \textbf{1.38} \\
\midrule

\multirow{2}{*}{CLIP Corr. $\uparrow$}
& MinD-Vis \cite{chen2022seeing} & \textbf{88.29} & - & - & 83.09 & -  \\
& \textit{Ours} & 86.60 & 83.56 & 83.82 & \textbf{84.77} & 84.69 \\
\midrule

\multirow{2}{*}{CLIP Dist. $\uparrow$}
& MinD-Vis \cite{chen2022seeing} & 0.41 & - & - & 0.37 & -  \\
& \textit{Ours} & \textbf{0.41} & 0.38 & 0.37 & \textbf{0.39} & 0.39 \\
\midrule

\multirow{2}{*}{Accuracy $\uparrow$}
& MinD-Vis \cite{chen2022seeing} & \textbf{29.94} & 18.50 & \textbf{21.00} & 20.37 & 22.45 \\
& \textit{Ours} & 28.48 & \textbf{19.74} & 19.65 & \textbf{22.48} & \textbf{22.59} \\
\bottomrule
\end{tabular}}}}
\end{table}

\subsection{Quantitative Results}
We compare the fMRI decoding performance on GOD and BOLD5000 datasets in Table~\ref{tab:god} and Table~\ref{tab:bold}, respectively.
Due to the availability of the official model of CIS-1,4 on the BOLD5000 dataset in MinD-Vis, we solely report the reproduced CLIP scores for these two individuals.
In both GOD and BOLD5000 datasets, our LEA enjoys superior performance compared with previous state-of-the-art algorithms on average in all image quality, semantic correctness and correlation perspectives.
Specifically, the generated image quality is consistently superior to competitors on GOD dataset.
While MinD-Vis has better performance on some subjects, it is significantly worse than LEA in other subjects, for example SUB-1,2,4 in GOD and CSI-2,4 in BOLD5000, yielding an inferior performance on average.
Hence the fMRI decoding performance of LEA is superior to current competitors while LEA also enjoys the capacity of fMRI encoding.

\begin{figure*}
\begin{center}
\centerline{\includegraphics[width=1\textwidth]{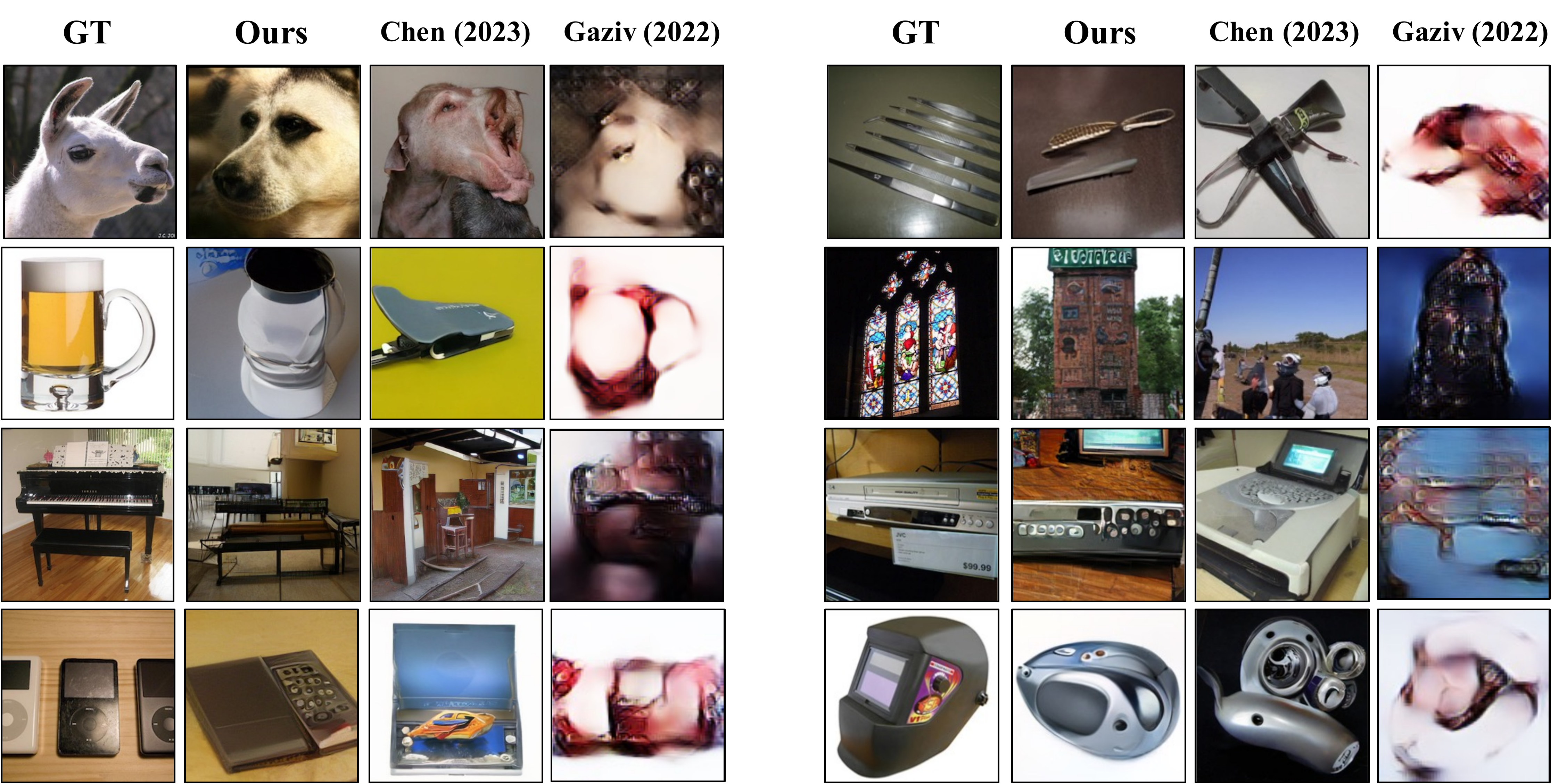}}
\vskip -0.1in
\caption{fMRI decoding performance comparison on GOD dataset. 
}
\label{fig:vis-god}
\end{center}
\vskip -0.2in
\end{figure*}

\begin{figure*}
\begin{center}
\centerline{\includegraphics[width=1\textwidth]{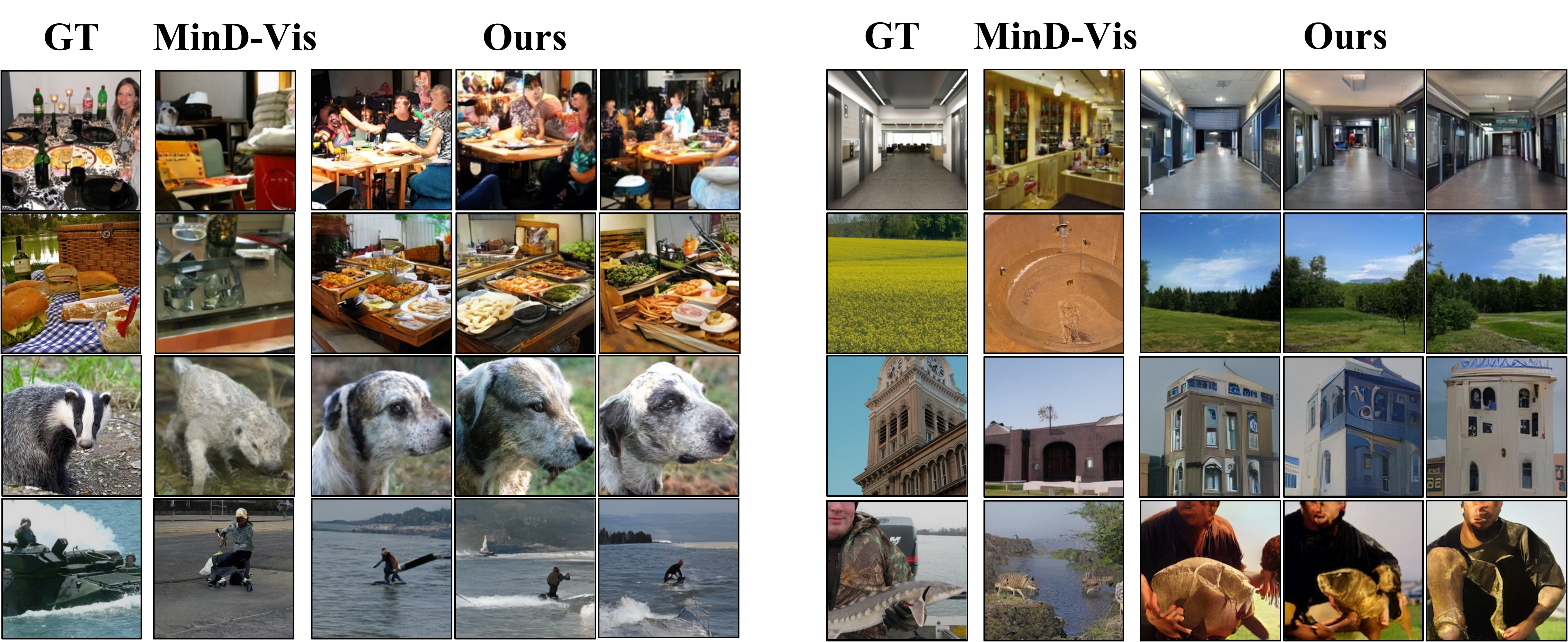}}
\vskip -0.1in
\caption{fMRI decoding performance comparison on BOLD5000 dataset. 
}
\label{fig:vis-bold}
\end{center}
\vskip -0.2in
\end{figure*}

\subsection{Qualitative Results}
To intuitively illustrate the efficacy of our method, we show some generated images in Figure~\ref{fig:vis-god} and Figure~\ref{fig:vis-bold} for GOD and BOLD5000 datasets, respectively. 
We compare with MinD-Vis~\cite{chen2022seeing} and Gaziv \textit{et al.}~\cite{gaziv2022self} on GOD and MinD-Vis on BOLD5000. 

\noindent \textbf{Our method can reconstruct images that are semantic-consistently with observed images.}
The generated images have the same semantic information as the ground-truth including the humans, objects, animals, architecture, and landscapes, while competitors may fail in some cases.
For example in Figure~\ref{fig:vis-god}, LEA  enjoys a semantic-consistent and high-fidelity generation, while competitors cannot preserve the semantics.
In the left last row of Figure~\ref{fig:vis-bold}, MinD-Vis can only generate a person and neglect the water, while LEA consistently generate both the person and the water.
Such results show that our generated images can not only retain the correct objects but also have more detailed semantics. 
Further, the fidelity of images is also achieved in many cases.

\begin{figure*}
\begin{center}
\centerline{\includegraphics[width=1\textwidth]{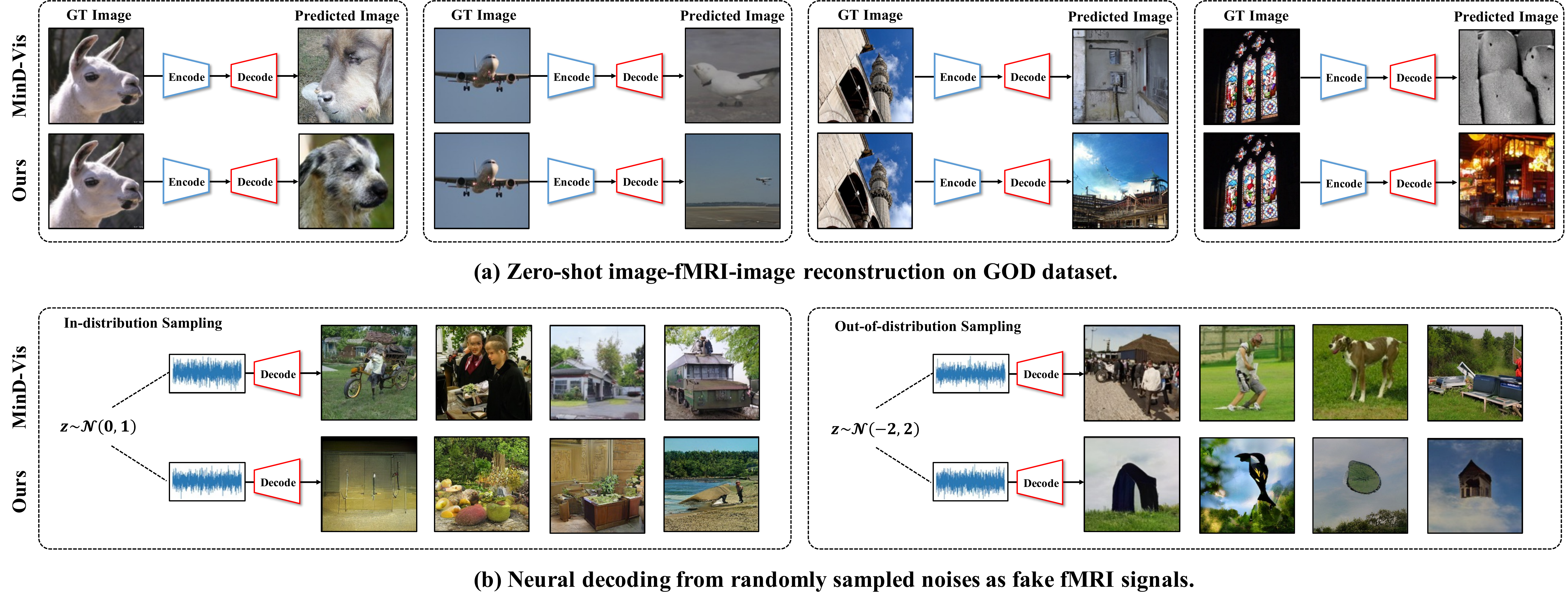}}
\vskip -0.15in
\caption{Reliable fMRI decoding Analysis.
}
\label{fig:img-fmri-img}
\end{center}
\vskip -0.2in
\end{figure*}

\subsection{Further Analysis}

\textbf{Reliable fMRI decoding}.
It is essential to ensure that the fMRI decoding model is accurately decoding the brain activities instead of simply performing unconditional generation. 
However, as current approaches usually adopt large-scale pre-trained generative models which can perform high-fidelity generation results even conditioned on random noise.
It is questionable if the generated high-fidelity image reflects the brain activities, especially when we perform zero-shot evaluations where the fMRI signals are unseen during training stage.

To check the reliability of fMRI decoding models, we conduct two experiments:
\textbf{(1)} Zero-shot image-fMRI-image reconstruction.
In this task, we use a novel image to estimate the fMRI signals, and subsequently use the estimated fMRI signals to reconstruct the image. 
The ability to reconstruct semantically consistent images serves as an indicator of the reliability of both the neural encoding and decoding models.
\textbf{(2)} fRMI-image with random noise as fake fMRI signals.
In this task, we randomly generate fMRI signals from Gaussian distribution that are far from the real fMRI signal distribution, and decode the fake fMRI signal to generate images.
The reconstructed image should contain non-semantic information to indicate a non-informative fMRI signals.

As illustrated in Figure~\ref{fig:img-fmri-img}, LEA works well on the image-fMRI-image reconstruction task, and generate unrealistic and non-informative images when conditioned on fake fMRI signals far from the distribution of real fMRI signals.
On the contrary, MinD-Vis perform poorly on the image-fMRI-image reconstruction task, and generates high-fidelity and informative images conditioned on random noise.
These findings indicate that existing models have not thoroughly explored the reliability of neural decoding models. However, in comparison, both the fMRI decoding and encoding components of LEA exhibit greater reliability than current algorithms.

\textbf{Limitation}. 
The primary limitation of LEA lies in its incomplete resolution of the instability issue associated with fMRI signals. We address this challenge by mitigating the problem through learning space compression, thereby necessitating only a linear model. This approach proves advantageous for constructing individual-specific models.
Ideally, the optimal solution would entail the development of an individual-independent model that operates consistently across all individuals.

\section{Conclusions} 
\label{sec:conclusions}
In this paper, we tackle the problem of joint fMRI decoding and encoding in a unified framework.
Our propose Latent Embedding Alignment (LEA) construct latent space for fMRI signals and images and align them to enable the bi-directional transformation.
While alleviating the redundancy, instability, and insufficiency issues of fMRI dataset, our LEA can produce high-fidelity semantic-consistent fMRI decoding results.
On the other hand, LEA can also perform fMRI encoding to direct estimate the human brain activity from visual stimuli.
Experiments on two benchmark datasets validate the effectiveness of LEA.

{\small
\bibliography{egbib}
\bibliographystyle{plain}
}

\end{document}